\documentclass[conference]{IEEEtran}
\usepackage{listings}
\usepackage[table,xcdraw]{xcolor}
\usepackage{threeparttable}
\usepackage[font=small,labelfont=bf,tableposition=top]{caption}
\DeclareCaptionLabelFormat{andtable}{#1~#2  \&  \tablename~\thetable}
\usepackage{amsmath}
\usepackage{pifont}
\usepackage{graphicx}
\usepackage{amssymb}
\definecolor{mydarkblue}{rgb}{0,0.08,0.65}
\usepackage[colorlinks=true,
    linkcolor=mydarkblue,
    citecolor=mydarkblue,
    filecolor=mydarkblue,
    urlcolor=mydarkblue]{hyperref} 
\usepackage[capitalize]{cleveref}
\usepackage{soul}
\usepackage[shortlabels]{enumitem}
\usepackage{url}
\usepackage{listings}
\usepackage{color}
\usepackage{tikz}
\usepackage{balance}
\usepackage{multirow}
\usepackage{comment}
\usepackage{amsmath}
\usepackage{graphicx}
\usepackage{wrapfig,lipsum,booktabs}
\usepackage[frozencache,cachedir=.]{minted}
\usepackage[symbol]{footmisc}
\usepackage{tablefootnote}

\usepackage{dblfloatfix}
\usepackage{natbib}

\usepackage{subfig}

\definecolor{codegreen}{rgb}{0,0.6,0}
\definecolor{codegray}{rgb}{0.5,0.5,0.5}
\definecolor{codepurple}{rgb}{0.58,0,0.82}
\definecolor{backcolour}{rgb}{0.95,0.95,0.92}

\usepackage[most]{tcolorbox}
\definecolor{sampleBg}{RGB}{250,250,250}
\definecolor{sampleBorder}{RGB}{200,200,200}
\definecolor{sampleLine}{RGB}{180,180,180}
\newtcolorbox{textbox}[1][]{
    enhanced,
    breakable,                 
    colback=sampleBg,         
    colframe=sampleBorder,    
    boxrule=0.5pt,           
    left=12pt,               
    right=8pt,              
    top=8pt,                
    bottom=8pt,             
    arc=0pt,                
    leftrule=2pt,           
    frame hidden,           
    overlay={
        \draw[sampleLine, line width=2pt] 
            ([xshift=3pt]frame.north west) -- 
            ([xshift=3pt]frame.south west);
    },
    fontupper=\ttfamily\small,   
    before skip=\medskipamount,  
    after skip=\medskipamount,   
    listing engine=listings,     
    listing options={
        basicstyle=\ttfamily\small,
        keepspaces=true,          
        columns=fullflexible,     
        breaklines=true,          
        breakatwhitespace=false,  
        showspaces=false,         
        showstringspaces=false,   
    },
    #1                           
}

\makeatletter
\def\blfootnote{\xdef\@thefnmark{}\@footnotetext}
\makeatother

\lstdefinestyle{mystyle}{
  backgroundcolor=\color{backcolour},   commentstyle=\color{codegreen},
  keywordstyle=\color{magenta},
  numberstyle=\tiny\color{codegray},
  stringstyle=\color{codepurple},
  basicstyle=\ttfamily\footnotesize,
  breakatwhitespace=false,         
  breaklines=true,                 
  captionpos=b,                    
  keepspaces=true,                 
  numbers=left,                    
  numbersep=5pt,                  
  showspaces=false,                
  showstringspaces=false,
  showtabs=false,                  
  tabsize=2,
}

\IEEEoverridecommandlockouts
\begin{document}

\makeatletter
  \def\title@font{\Large}
  \let\ltx@maketitle\@maketitle
  \def\@maketitle{\bgroup%
    \let\ltx@title\@title%
    \def\@title{\resizebox{\textwidth}{!}{%
      \mbox{\title@font\ltx@title}%
    }}%
    \ltx@maketitle%
  \egroup}
\makeatother

\title{Zyda-2: a 5 Trillion Token High-Quality Dataset}
\author{Yury Tokpanov $\quad$ Paolo Glorioso $\quad$ Quentin Anthony $\quad$ Beren Millidge \\
{
\small
\{yury, paolo, quentin beren\}@zyphra.com
}\\

{}\\
{
 Zyphra, Palo Alto, CA
 \small
}\\

}

\maketitle

\setcounter{page}{1}

\begin{abstract}

In this technical report, we present Zyda-2: a five trillion token dataset for language model pretraining. Zyda-2 was used to train our Zamba2 series of models which are state-of-the-art for their weight class. We build Zyda-2 by collating high-quality open-source tokens such as FineWeb and DCLM, then distilling them to the highest-quality subset via cross-deduplication and model-based quality filtering. Zyda-2 is released under a permissive open license, and is available at \url{https://huggingface.co/datasets/Zyphra/Zyda-2}.

\end{abstract}

\section{Introduction}

Two of the primary determinants of the quality of a language model are the quality and scale of the datasets used to train it.
For data quality, initial models such as GPT3~\citep{gpt3} were trained purely on web datasets coarsely filtered from Common Crawl\footnote{\url{https://commoncrawl.org/}}. From there, improving quality via diversity of data sources and the importance of filtering and deduplicating sources have been championed by datasets such as The Pile \citep{gao2020pile} and FineWeb \citep{penedo2024fineweb}, respectively.
The scale of datasets used for pretraining has also increased dramatically: while GPT3~\citep{gpt3} was trained on only a few hundred billion tokens, current much-smaller models are trained on trillions of tokens. One example of this is Llama3-8b \citep{llama3}, which was trained on 15 trillion tokens. 
For open-source models to remain on par with closed ones, open-source datasets also need to remain competitive in both quality and scale. To meet this need for open-source, large-scale, and high-quality datasets, sophisticated data pipelines have emerged which scrape, filter, deduplicate, and mix data sources.

Open-source datasets have been a key driver of model quality, with C4 \citep{raffel2020exploring} and the Pile \citep{gao2020pile} being the first open language modeling dataset of sufficient size to serve as a pretraining dataset. More recent datasets such as Dolma \citep{soldaini2024dolma} and RefinedWeb \citep{penedo2023refinedweb} have added additional levels of syntactic filtering and deduplication, raising their quality. A more recent trend has been the use of small classifier models to perform syntactic filtering. For instance, filtering for some proxy of `educational content' has been used to great effect in the Phi \citep{abdin2024phi} series of models which---although most details are not published---appear to utilize both heavy filtering and a large amount of synthetic data generation. Open datasets utilizing this model-based filtering approach include FineWeb-Edu \citep{penedo2024fineweb} and DCLM \citep{li2024dclm}, which perform extremely well compared to prior datasets. 

At Zyphra, one of our key priorities is producing the highest-quality and most efficient models for a given parameter budget, with a special focus on small, highly powerful models which can automate many tasks cheaply and be utilized on consumer and edge devices. This has driven our innovations in model architecture \citep{glorioso2024zamba,anthony2024blackmamba} and also necessitates a focus on constructing strong pretraining and annealing datasets in order to maximize the performance per FLOP and per parameter during training. High-quality datasets appear especially important for smaller models, since they have less total capacity and hence are more affected by significant quantities of noise or low-quality tokens in their training datasets.

Our general approach to dataset creation is to collect all openly available and highly-performing open-source datasets and improve their quality further by removing duplicates and adding additional filtering steps. We then weight the resulting dataset mixture to obtain highest-quality subset that meets our training budget. Our previous dataset constructed with this approach was Zyda-1 \citep{tokpanov2024zyda}, which was used to train Zamba1-7B \citep{glorioso2024zamba}. Zyda-1 outperformed all major language modeling datasets at the time, such as Dolma-1.6~\cite{soldaini2024dolma}, FineWeb~\citep{penedo2024fineweb}, and RefinedWeb\citep{penedo2023refinedweb}. Since then, however, new datasets utilizing model-based filtering have been released which gave a significant boost to performance and eclipsed Zyda-1. For Zyda-2, we aimed to significantly increase the scale and quality of the dataset beyond Zyda-1, to enable us to reach the frontier of performance for small language models. 

Zyda-2 was used to train our Zamba2 series of models \citep{glorioso2024zamba2} that achieve state-of-the-art performance at all of the 1.2B, 2.7B, and 7B parameter brackets, beating strong comparable models such as Meta's Llama3 series \citep{llama3} and Google's Gemma series \citep{team2024gemma}. As such, Zyda-2 provides important information about what kind of pretraining dataset is necessary to reach the performance frontier at present.

Zyda-2 is released under a permissive open-source license (ODC-BY) and can be found on HuggingFace at \url{https://huggingface.co/datasets/Zyphra/Zyda-2}

\section{Dataset Construction}

\begin{figure}
    \centering
    \includegraphics[width=0.9\linewidth]{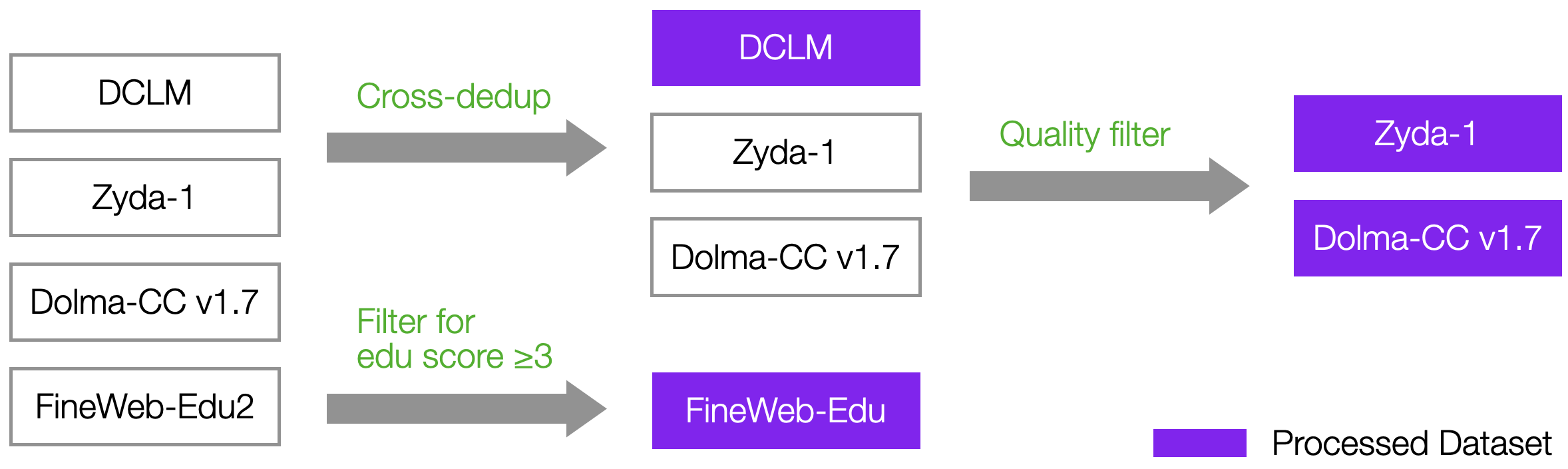}
    \caption{Dataset creation process for Zyda-2. We first collated the best open-source sets available, then ran cross-deduplication between these datasets, since they all ultimately derive mostly from a common source (common-crawl). Finally, we applied model-based quality filtering to the two unfiltered datasets (Zyda-1 and Dolma-CC).}
    \label{dataset_construction}
\end{figure}

Zyda-2 was built upon the following data sources: DCLM-baseline-1.0 (DCLM for short), FineWeb-Edu-score-2 (FineWeb-Edu2 for short), Zyda-1 and the Common Crawl portion of Dolma v1.7 (Dolma-CC for short)\footnote{The other portions of Dolma are separate math, code, and instruct datasets which can be mixed in independently or in an annealing phase.}. We chose the FineWeb-Edu2 version of FineWeb-Edu in order to have a much bigger starting pool of documents. These datasets were put through a two-stage pipeline: a cross-deduplication phase between all datasets, followed by a model-based filtering stage applied to the Zyda-1 and Dolma-CC datasets (see Fig. \ref{dataset_construction}).

\subsection{Cross-Deduplication}

Given that all open datasets ultimately originated from similar Common Crawl scrapes, we expect them to contain significant fractions of duplicated documents. Deduplication has been found to generally improve language modeling performance \citep{lee2021deduplicating}, although recent papers have claimed its effects can be neutral and perhaps negative when applied in the wrong context \citep{li2024dclm}. 

We used approximate minhash LSH deduplication for our deduplication pipeline with the following parameters: minhash with signature size of 128 computed on character-based 25-grams signatures and split into 8 bands, giving roughly 85\% Jaccard similarity threshold. We then constructed an undirected graph with nodes being documents and edges being duplicates, and found connected components in it, which provided us with clusters of duplicates. From each cluster, we selected the top document to keep and removed the rest. We selected the remaining document according its origin in the following ranking of datasets: FineWeb-Edu2 $>$ DCLM $>$ Zyda-1 $>$ Dolma-CC; that is, when choosing a document to keep, we kept the one from the highest-ranked dataset.

As expected, we found a significant number of duplicated documents across datasets, which resulted in the removal of approximately 11\% of the total tokens (or 13\% of total documents) from Zyda-2 compared to its component datasets. 

Additionally, we performed a full intra-dataset deduplication of both DCLM and FineWeb-Edu, since we found that both datasets contained a very large number of internal duplicates (about 80\% for both datasets). Both the DCLM and FineWeb-Edu papers claim that internal dataset deduplication did not improve their results and so they did not pursue full deduplication or release a deduplicated version of their datasets. In our initial ablation experiments, we partly confirmed their results that this deduplication had only a small effect. This raises the future research questions of: 1) When is duplication harmful and when is it not? 2) Why are models seemingly robust to being trained on duplicate data?

\subsection{Model-based filtering}

The second phase of processing was model-based filtering. We only applied model-based filtering to Zyda-1 and Dolma-CC. This is because Zyda-1 and Dolma-CC are less filtered and contain a higher variety of internet documents which are not all designed to be educational. DCLM and FineWeb-Edu, however, have already undergone significant quality filtering as a core component of their creation, and indeed we did not observe benefits in performing additional filtering from our training ablations.

\begin{figure}
    \centering
    \includegraphics[width=\linewidth]{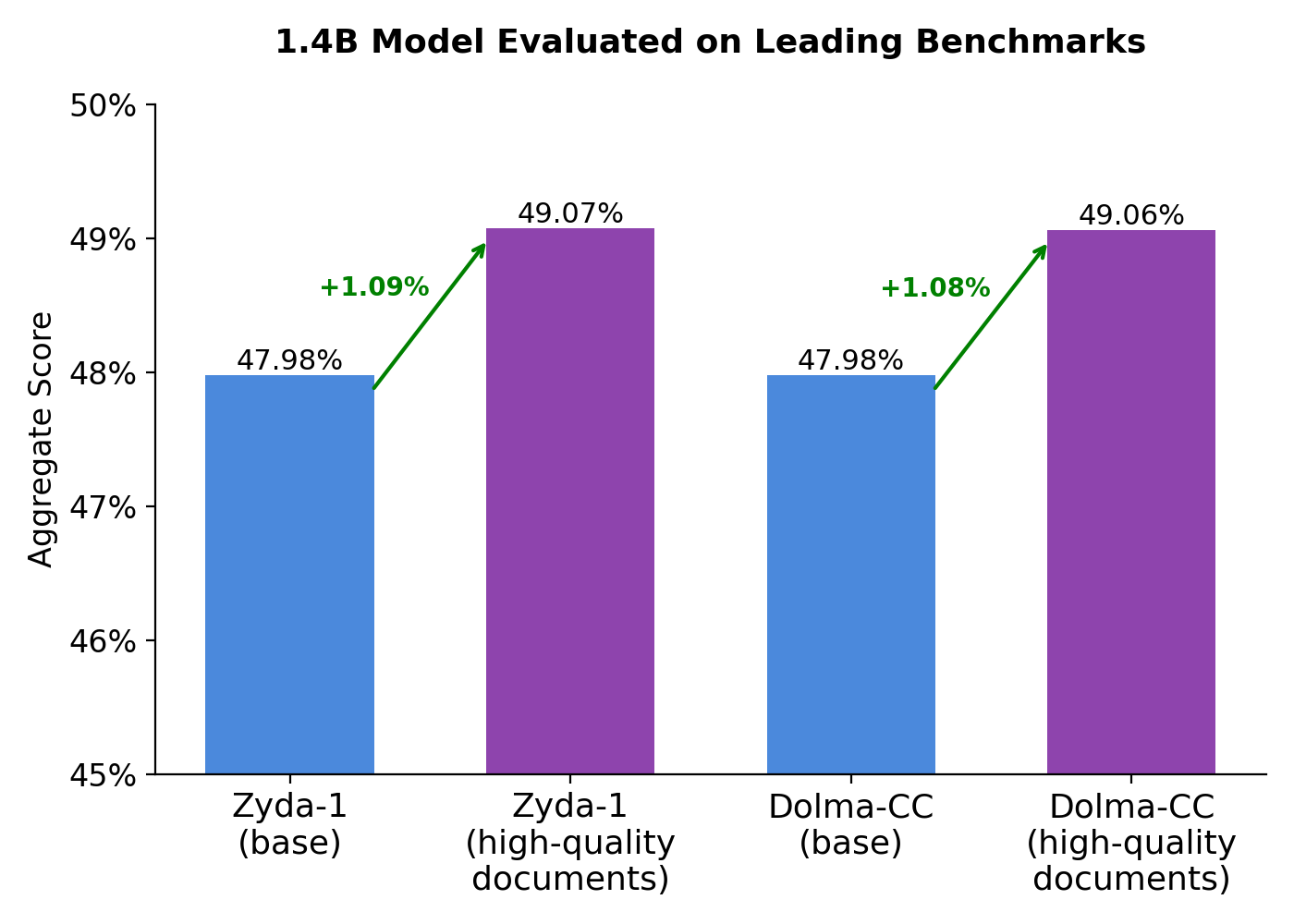}
    \caption{The performance of a 1.4B model trained on 50B tokens with and without model-based filtering on the Zyda-1 and Dolma-CC datasets. The aggregate evaluation score is the mean across the following standard language modeling benchmarks: Hellaswag, PIQA, OpenBookQA, Arc-Challenge, Arc-Easy, and Winogrande. For the quality filtering we kept only those documents labeled as `high-quality' by the model-based classifier.}
    \label{fig:model_filtering_ablation}
\end{figure}

For this step, we experimented with the quality-classifier-deberta model provided in NeMo Curator \citep{nemo_curator}. We applied this model to filter both Zyda-1 and Dolma-CC, and experimented with either removing only the ‘low’ quality documents or keeping only the ‘high’ quality ones. In an ablation study where we trained a 1.4B parameter transformer for 50B tokens, we found that keeping only the highest quality 10-20\% of the documents significantly improved model performance for both Zyda-1 and Dolma-CC (see Figure \ref{fig:model_filtering_ablation}), while only removing ‘low’ quality documents had less effect.

This performance improvement likely occurs because these datasets are not pre-filtered by similar classifiers and demonstrates that, even though both datasets have undergone thorough syntactic filtering, significant gains can be had by using model-based quality classifiers. However, re-filtering already filtered datasets such as DCLM provided a negligible gain.

As an additional step, we filtered FineWeb-Edu2 by its educational score. This step essentially converts it to FineWeb-Edu, a higher-quality subset of FineWeb-Edu2 comprising samples with the higher score 3 based on the FineWeb-Edu quality filter model. This scoring method yielded the best performance in the ablations conducted during the creation of FineWeb-Edu. Notably, we applied this filtering after cross-deduplicating DCLM against FineWeb-Edu2, which involved eliminating all samples in DCLM that were deemed duplicates with FineWeb-Edu2. This sequence of steps indirectly applied the FineWeb-Edu quality classifier to DCLM, effectively removing the samples in DCLM with lower educational content.


\subsection{Composition}

\begin{figure}
    \centering
    \includegraphics[width=0.9\linewidth]{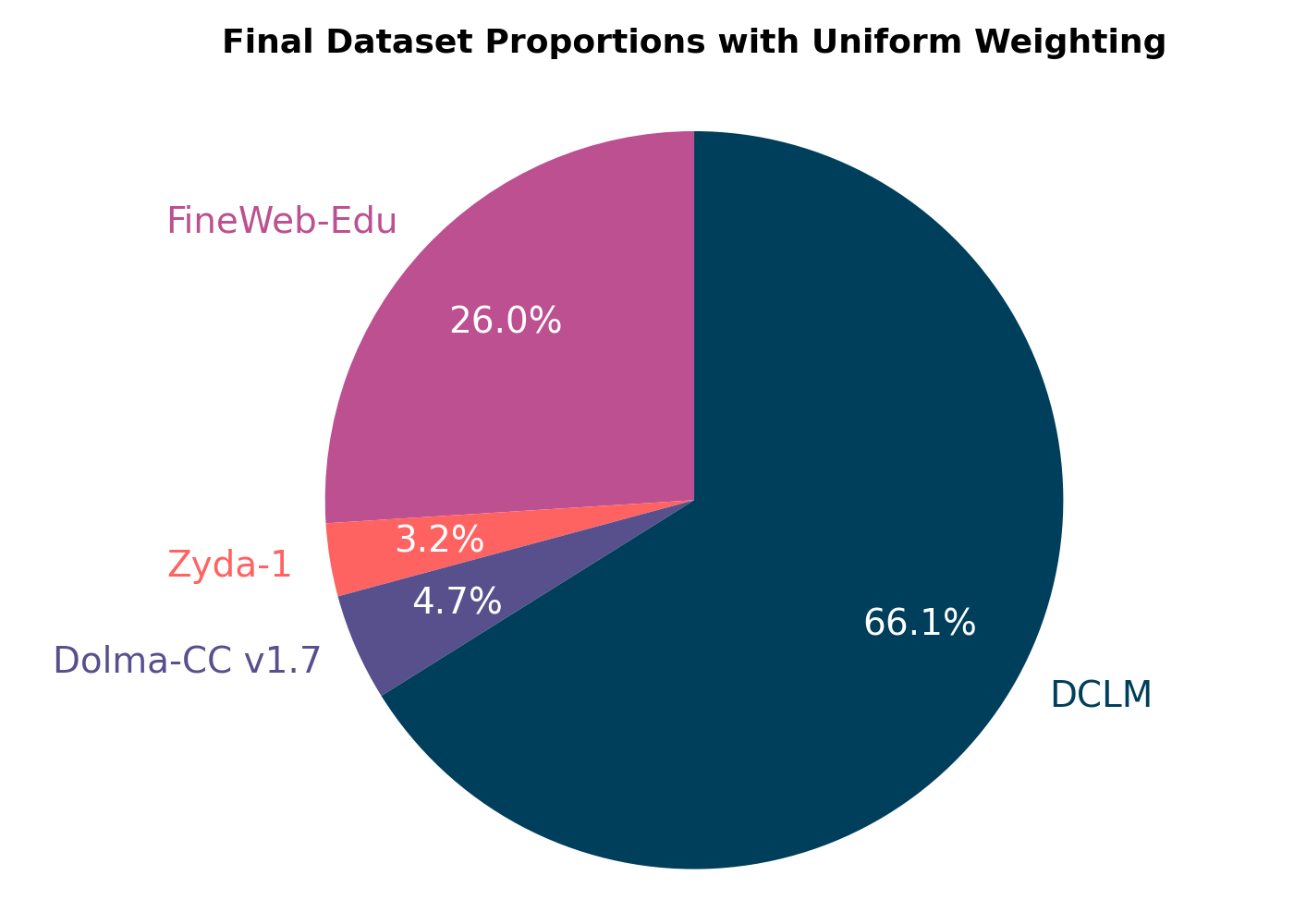}
    \caption{Composition of Zyda-2}
  \label{fig:uniform_weighting_pie}
\end{figure}

Following the dataset processing steps that we described above resulted in a total dataset consisting of approximately five trillion tokens. The number of tokens removed in each step of our pipeline is presented in \cref{table:table_dataset_stages}. Additionally, the fraction of the total dataset comprised by each of the component datasets is presented in Figure \ref{fig:uniform_weighting_pie}.

\begin{table}[h]
\begin{center}
\begin{threeparttable}
\caption{The number of tokens (GPT-Neox tokenizer) in Zyda-2 at each step of processing, from the initial dataset collection through cross-deduplication and then model-based filtering. The number of tokens in FW-Edu stays fixed because it is treated as the `highest-rank' deduplication dataset.}\label{table:table_dataset_stages}
\begin{tabular}{l c c c}
\toprule
\multicolumn{4}{c}{\textbf{Dataset Statistics} (in Billion Tokens)} \\
\hline
& Start & Cross-Dedup & Filter \\
\hline
DCLM            & 3.850 & 3.348 & 3.348 \\
Dolma-CC v1.7   & 1.209 & 0.969 & 0.238 \\
Zyda-1          & 1.056 & 0.937 & 0.163 \\
FineWeb-Edu     & 1.319 & 1.319 & 1.319 \\
\midrule
\midrule
Total & 7.434 & 6.573 & 5.068 \\
\bottomrule
\end{tabular}
\end{threeparttable}
\end{center}
\end{table}

The majority of documents come from DCLM, which is expected given its large size compared to the other datasets. The second-largest dataset is FineWeb-Edu which comprises approximately one quarter of the dataset, with Zyda-1 and Dolma-CC having smaller contributions.

\section{Performance}

\begin{figure}[t]
    \centering
    \includegraphics[width=1.1\linewidth]{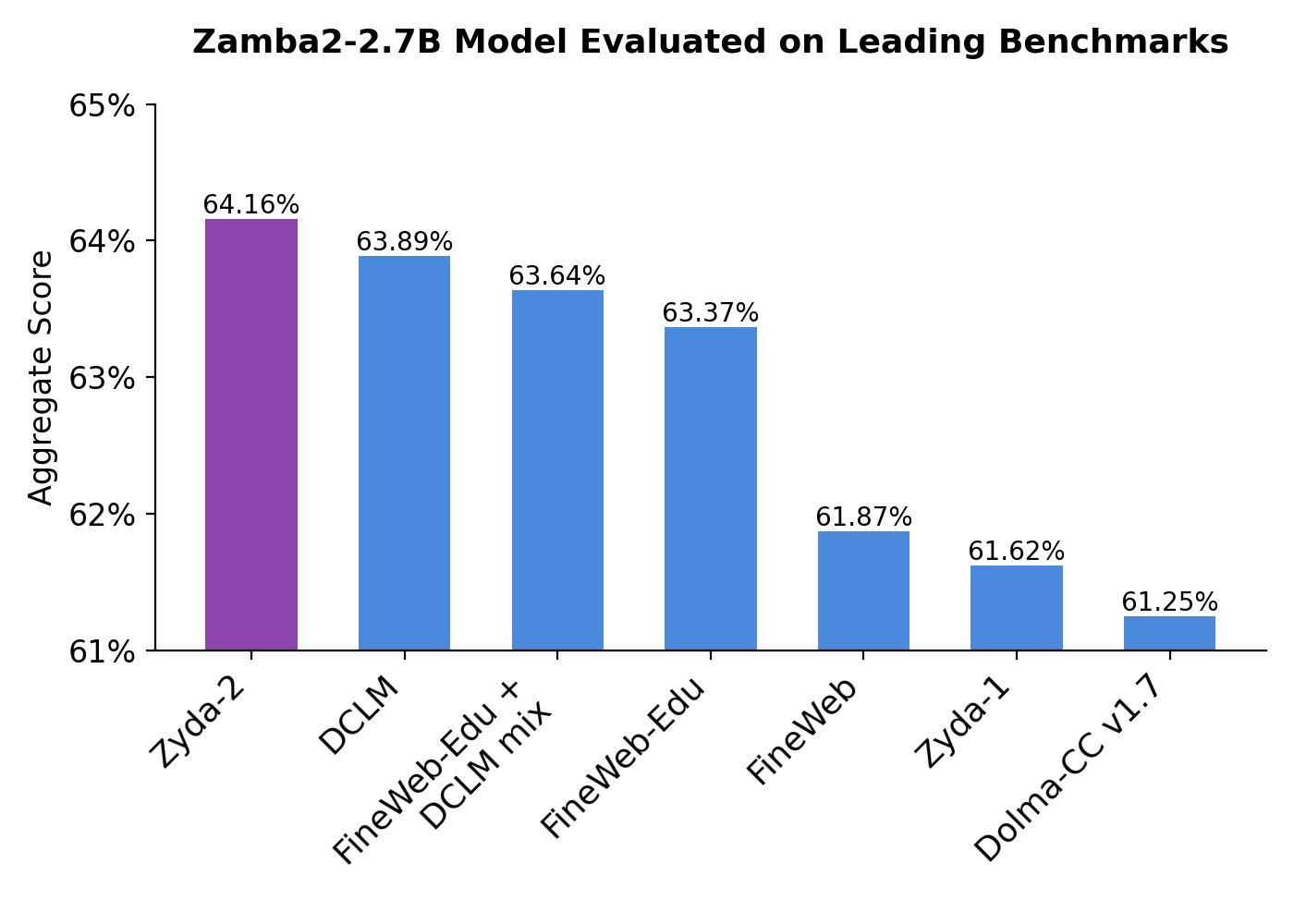}
    \caption{Performance of Zyda-2 vs other datasets as aggregate weighted evaluation score. This score is an average of MMLU, Hellaswag, PIQA, OpenBookQA, Arc-Challenge, Arc-Easy, and Winogrande. These scores are collected by annealing the base version of Zamba2-2.7B for roughly 40B tokens on each dataset.}
    \label{fig:Zyda-2-perf}
\end{figure}

\begin{figure}[t]
    \centering
    \includegraphics[width=0.9\linewidth]{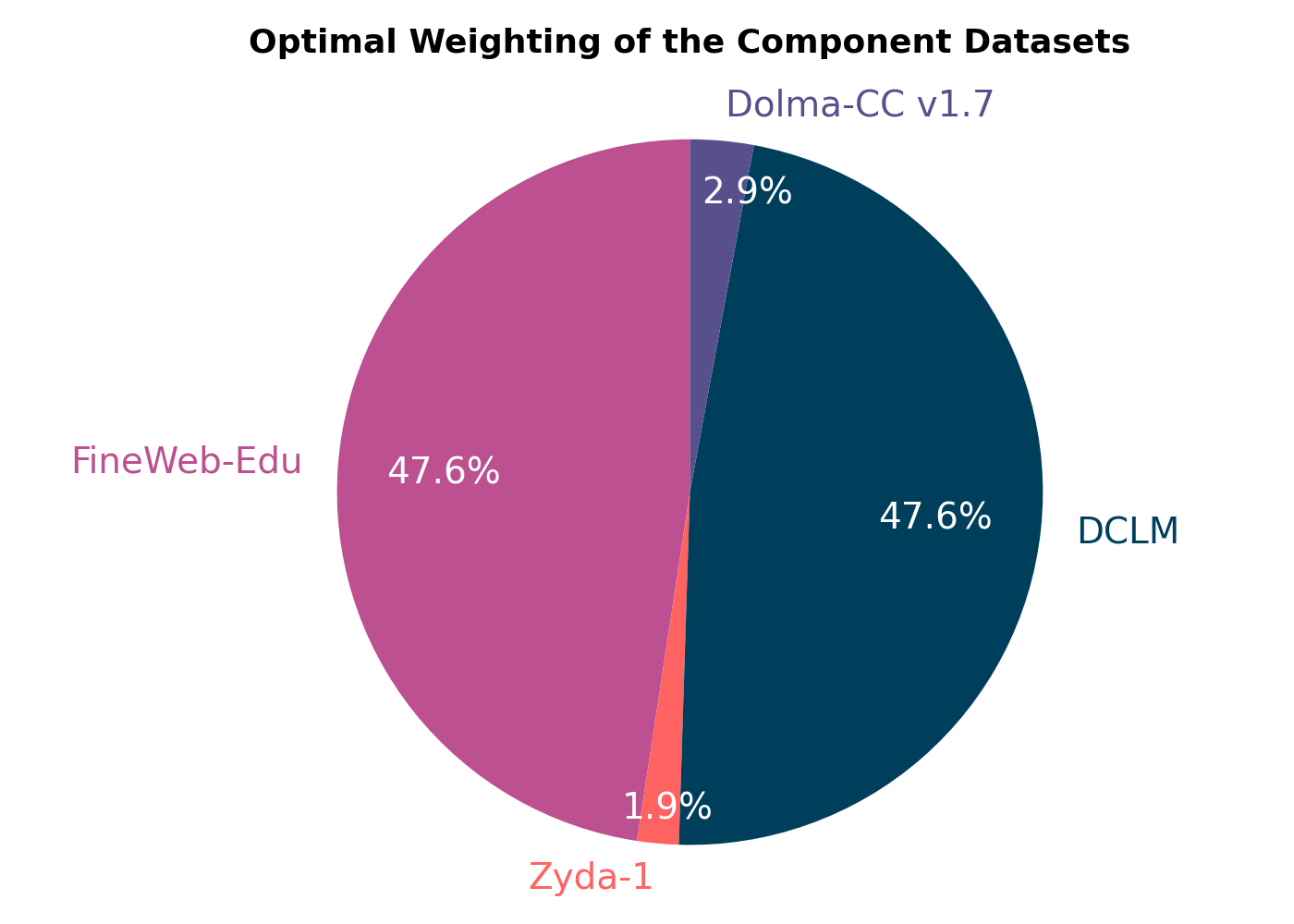}
    \caption{The proportion of each of the component datasets comprising Zyda-2 using the optimal weighting. FineWeb-Edu and DCLM each account for approximately the same total proportion of the dataset.}
    \label{optimal_weighted_pie_chart}
\end{figure}

Rigorously evaluating the quality of datasets is challenging without training large-scale models upon them, which costs significant compute per experiment. One typical alternative to get a reliable signal of dataset quality is to train small models on small slices of the dataset (for instance a 300 million-parameter model on 50 billion tokens) and then compare the evaluation performance of such models. This approach is taken by the majority of recent datasets released. This approach has advantages in that it is now computationally feasible to run many ablation experiments even on relatively constrained resources. However, many standard evaluations do not show strong signal for small models and low token counts (for instance, MMLU remains at chance for such models). 

An alternative approach is to utilize the newly popular annealing approach to training LLMs \citep{hu2024minicpm,parmar2024nemotron,glorioso2024zamba}, which involves training an LLM on a large amount of data with a relatively slow learning-rate decay followed by a rapid learning rate decay over a high quality dataset. \citet{blakeney2024does} proposes utilizing the annealing phase as a method to test the performance of different datasets. While not requiring significantly more compute than training a small model from scratch, annealing enables larger and already trained base models to be used which can show signal on important evaluation metrics such as MMLU. We follow this methodology and indeed observe significantly more signal on our dataset ablations when testing with annealing rather than training small models from scratch. We perform our annealing ablations on our pre-annealing checkpoint of our Zamba2-2.7B model \cite{glorioso2024zamba2} which we run for 40 billion annealing tokens. Since our pre-annealing Zamba2-2.7B model is significantly above chance at MMLU and is already highly trained on approximately 3 trillion token, we observe a clear signal on all standard ablations and additionally greater sensitivity in evaluation scores to changes in dataset composition. We observe that Zyda-2 outperforms the currently leading datasets in aggregate evaluation scores and performs significantly better than any of its component datasets. This is due both to our additional filtering and deduplication pipeline and also an ensembling effect whereby adding in more varied data sources improves overall performance and robustness versus any one specific dataset.

\subsection{Subset Weightings}

To boost performance further, we investigated the optimal weightings of Zyda-2's component datasets. We conducted a series of experiments to determine this using our annealing scheme described previously. We found that a uniform weighting scheme, where each dataset is weighted by its total number of tokens, is suboptimal. Instead, upweighting FineWeb-Edu such that it became of equal proportion to DCLM exceeded the performance of the uniform weighting approach. With this weighting, due to their smaller sizes, Zyda-1 and Dolma-CC in total make up approximately 5\% of the total dataset. However, we found that removing Zyda-1 and Dolma-CC entirely worsened performance, demonstrating that although their total token count is small, adding these datasets brings much-needed diversity of sources to the Zyda-2 dataset.

\section{Discussion}

In this paper we have presented the Zyda-2 dataset, which achieves leading performance across many language evaluations. At 5 trillion tokens, Zyda-2 is perfect for large-scale pretraining. Zyda-2 is primarily focused on natural language capabilities, and, for a generalist language model, we recommend augmenting it with specialized datasets especially for coding, mathematical reasoning, and any other niche capabilities required. We constructed Zyda-2 by collecting the best existing open datasets and applying a two-stage process of cross-deduplication and model-based filtering. We found that model-based filtering significantly improved the performance of the existing unfiltered datasets while having little effect on the already filtered datasets, showing that performing additional educational quality filtering on an already-filtered dataset is not helpful. 

While performing deduplication of DCLM and FineWeb-Edu, we found that both datasets contained large fractions of internal duplicates. This result raises several interesting questions. First, it is unclear why removing duplicates does not improve performance. This implies that documents with large numbers of identical tokens provide approximately the same benefit as fresh tokens from the same distribution. We can perhaps consider a highly duplicated dataset like FineWeb-Edu as equivalent to performing a 2-5 epoch shuffled run on a much smaller deduplicated dataset, implying that a small number of epochs do not particularly harm language model performance at these scales. However, this effect may not hold at larger scales, where language models are more sample-efficient at memorization of the training dataset, thus requiring more regularization to counter overfitting. On the other hand, an argument can be made that samples that are repeated more may be of higher quality, although having looked at such highly repetitive samples we doubt this since most seem to be related to common preambles or other features of websites. If this is true, however, it is unclear why earlier results showed positive effects of deduplication \citep{lee2021deduplicating}. One possible hypothesis is that the early datasets where deduplication improved performance were not strongly filtered by model-based classifiers like DCLM and FineWeb-Edu, and thus the deduplication step may simply be removing many low-quality spam documents that occur often in unfiltered web data. In more stringently filtered datasets, the magnitude of this effect could diminish. However, we do note that large clusters of duplicates in DCLM still contain low-quality documents: the most frequent duplicate in DCLM is indeed a spam message. Additionally, we notice that on supposedly duplicate data, the scores assigned by the quality-filtering model often vary significantly, despite only very minor differences in text. This is perhaps indicative of the relative lack of robustness in some of these classifiers or the presence of significant fractions of false-positives in the deduplication step.

Model-based quality filtering based on `educational quality' as performed by FineWeb-Edu and DCLM has created a marked increase in language modeling performance, at least on commonly used language modeling benchmarks such as MMLU and the ARC tasks, which is where we observe the largest increases. Interesting questions remain around the degree to which model-based filtering can be further improved beyond current methods. Synthetic data \citep{abdin2024phi}, including augmentations of existing datasets \citep{maini2024rephrasing}, shows considerable promise \citep{maini2024rephrasing}, as does methods for using other filtering approaches such as language modeling perplexity \citep{ankner2024perplexed}.

\clearpage

\section*{Author Contributions}

\textbf{Yury} — Project lead. Performed majority of implementation and obtained experimental results. Primary coordinator with Nvidia.

\textbf{Paolo} — Contributed to model-based-filtering and deduplication. 

\textbf{Quentin} — Contributed to model-based-filtering and deduplication. Contributed to dataset conceptualization. Secondary coordinator with Nvidia.

\textbf{Beren} — Overall project supervisor. Contributed to dataset conceptualization. Contributed to model-based-filtering and deduplication.  Primary writer of technical report.

\section*{Acknowledgements}

We would like to acknowledge Anna Golubeva's very helpful comments and edits on the draft of this paper. We would also like to acknowledge NVIDIA for their technical assistance and support with our use of their NeMoCurator library.

\bibliographystyle{apalike}
\bibliography{main}

\begin{thebibliography}{}

\bibitem[Abdin et~al., 2024]{abdin2024phi}
Abdin, M., Jacobs, S.~A., Awan, A.~A., Aneja, J., Awadallah, A., Awadalla, H., Bach, N., Bahree, A., Bakhtiari, A., Behl, H., et~al. (2024).
\newblock Phi-3 technical report: A highly capable language model locally on your phone.
\newblock {\em arXiv preprint arXiv:2404.14219}.

\bibitem[Ankner et~al., 2024]{ankner2024perplexed}
Ankner, Z., Blakeney, C., Sreenivasan, K., Marion, M., Leavitt, M.~L., and Paul, M. (2024).
\newblock Perplexed by perplexity: Perplexity-based data pruning with small reference models.
\newblock {\em arXiv preprint arXiv:2405.20541}.

\bibitem[Anthony et~al., 2024]{anthony2024blackmamba}
Anthony, Q., Tokpanov, Y., Glorioso, P., and Millidge, B. (2024).
\newblock Blackmamba: Mixture of experts for state-space models.
\newblock {\em arXiv preprint arXiv:2402.01771}.

\bibitem[Blakeney et~al., 2024]{blakeney2024does}
Blakeney, C., Paul, M., Larsen, B.~W., Owen, S., and Frankle, J. (2024).
\newblock Does your data spark joy? performance gains from domain upsampling at the end of training.
\newblock {\em arXiv preprint arXiv:2406.03476}.

\bibitem[Brown et~al., 2020]{gpt3}
Brown, T.~B., Mann, B., Ryder, N., Subbiah, M., Kaplan, J., Dhariwal, P., Neelakantan, A., Shyam, P., Sastry, G., Askell, A., Agarwal, S., Herbert{-}Voss, A., Krueger, G., Henighan, T., Child, R., Ramesh, A., Ziegler, D.~M., Wu, J., Winter, C., Hesse, C., Chen, M., Sigler, E., Litwin, M., Gray, S., Chess, B., Clark, J., Berner, C., McCandlish, S., Radford, A., Sutskever, I., and Amodei, D. (2020).
\newblock Language models are few-shot learners.
\newblock {\em CoRR}, abs/2005.14165.

\bibitem[Gao et~al., 2020]{gao2020pile}
Gao, L., Biderman, S., Black, S., Golding, L., Hoppe, T., Foster, C., Phang, J., He, H., Thite, A., Nabeshima, N., et~al. (2020).
\newblock The pile: An 800gb dataset of diverse text for language modeling.
\newblock {\em arXiv preprint arXiv:2101.00027}.

\bibitem[{Gemma Team} et~al., 2024]{team2024gemma}
{Gemma Team}, Mesnard, T., Hardin, C., Dadashi, R., Bhupatiraju, S., Pathak, S., Sifre, L., Rivi{\`e}re, M., Kale, M.~S., Love, J., et~al. (2024).
\newblock Gemma: Open models based on gemini research and technology.
\newblock {\em arXiv preprint arXiv:2403.08295}.

\bibitem[Glorioso, 2024]{glorioso2024zamba2}
Glorioso, P. (2024).
\newblock Zamba2.

\bibitem[Glorioso et~al., 2024]{glorioso2024zamba}
Glorioso, P., Anthony, Q., Tokpanov, Y., Whittington, J., Pilault, J., Ibrahim, A., and Millidge, B. (2024).
\newblock Zamba: A compact 7b ssm hybrid model.

\bibitem[Hu et~al., 2024]{hu2024minicpm}
Hu, S., Tu, Y., Han, X., He, C., Cui, G., Long, X., Zheng, Z., Fang, Y., Huang, Y., Zhao, W., et~al. (2024).
\newblock Minicpm: Unveiling the potential of small language models with scalable training strategies.
\newblock {\em arXiv preprint arXiv:2404.06395}.

\bibitem[Lee et~al., 2021]{lee2021deduplicating}
Lee, K., Ippolito, D., Nystrom, A., Zhang, C., Eck, D., Callison-Burch, C., and Carlini, N. (2021).
\newblock Deduplicating training data makes language models better.
\newblock {\em arXiv preprint arXiv:2107.06499}.

\bibitem[Li et~al., 2024]{li2024dclm}
Li, J., Fang, A., Smyrnis, G., Ivgi, M., Jordan, M., Gadre, S., Bansal, H., Guha, E., Keh, S., Arora, K., Garg, S., Xin, R., Muennighoff, N., Heckel, R., Mercat, J., Chen, M., Gururangan, S., Wortsman, M., Albalak, A., Bitton, Y., Nezhurina, M., Abbas, A., Hsieh, C.-Y., Ghosh, D., Gardner, J., Kilian, M., Zhang, H., Shao, R., Pratt, S., Sanyal, S., Ilharco, G., Daras, G., Marathe, K., Gokaslan, A., Zhang, J., Chandu, K., Nguyen, T., Vasiljevic, I., Kakade, S., Song, S., Sanghavi, S., Faghri, F., Oh, S., Zettlemoyer, L., Lo, K., El-Nouby, A., Pouransari, H., Toshev, A., Wang, S., Groeneveld, D., Soldaini, L., Koh, P.~W., Jitsev, J., Kollar, T., Dimakis, A.~G., Carmon, Y., Dave, A., Schmidt, L., and Shankar, V. (2024).
\newblock Datacomp-lm: In search of the next generation of training sets for language models.

\bibitem[Maini et~al., 2024]{maini2024rephrasing}
Maini, P., Seto, S., Bai, H., Grangier, D., Zhang, Y., and Jaitly, N. (2024).
\newblock Rephrasing the web: A recipe for compute and data-efficient language modeling.
\newblock {\em arXiv preprint arXiv:2401.16380}.

\bibitem[Meta, 2024]{llama3}
Meta (2024).
\newblock {Introducing Meta Llama 3: The most capable openly available LLM to date}.
\newblock https://ai.meta.com/blog/meta-llama-3/.
\newblock {Accessed: \today}.

\bibitem[Nvidia, 2024]{nemo_curator}
Nvidia (2024).
\newblock Nemo-curator: a toolkit for data curation.

\bibitem[Parmar et~al., 2024]{parmar2024nemotron}
Parmar, J., Prabhumoye, S., Jennings, J., Patwary, M., Subramanian, S., Su, D., Zhu, C., Narayanan, D., Jhunjhunwala, A., Dattagupta, A., et~al. (2024).
\newblock Nemotron-4 15b technical report.
\newblock {\em arXiv preprint arXiv:2402.16819}.

\bibitem[Penedo et~al., 2024]{penedo2024fineweb}
Penedo, G., Kydlíček, H., allal, L.~B., Lozhkov, A., Mitchell, M., Raffel, C., Werra, L.~V., and Wolf, T. (2024).
\newblock The fineweb datasets: Decanting the web for the finest text data at scale.

\bibitem[Penedo et~al., 2023]{penedo2023refinedweb}
Penedo, G., Malartic, Q., Hesslow, D., Cojocaru, R., Cappelli, A., Alobeidli, H., Pannier, B., Almazrouei, E., and Launay, J. (2023).
\newblock The refinedweb dataset for falcon llm: outperforming curated corpora with web data, and web data only.
\newblock {\em arXiv preprint arXiv:2306.01116}.

\bibitem[Raffel et~al., 2020]{raffel2020exploring}
Raffel, C., Shazeer, N., Roberts, A., Lee, K., Narang, S., Matena, M., Zhou, Y., Li, W., and Liu, P.~J. (2020).
\newblock Exploring the limits of transfer learning with a unified text-to-text transformer.
\newblock {\em Journal of machine learning research}, 21(140):1--67.

\bibitem[Soldaini et~al., 2024]{soldaini2024dolma}
Soldaini, L., Kinney, R., Bhagia, A., Schwenk, D., Atkinson, D., Authur, R., Bogin, B., Chandu, K., Dumas, J., Elazar, Y., Hofmann, V., Jha, A.~H., Kumar, S., Lucy, L., Lyu, X., Lambert, N., Magnusson, I., Morrison, J., Muennighoff, N., Naik, A., Nam, C., Peters, M.~E., Ravichander, A., Richardson, K., Shen, Z., Strubell, E., Subramani, N., Tafjord, O., Walsh, P., Zettlemoyer, L., Smith, N.~A., Hajishirzi, H., Beltagy, I., Groeneveld, D., Dodge, J., and Lo, K. (2024).
\newblock Dolma: an open corpus of three trillion tokens for language model pretraining research.

\bibitem[Tokpanov et~al., 2024]{tokpanov2024zyda}
Tokpanov, Y., Millidge, B., Glorioso, P., Pilault, J., Ibrahim, A., Whittington, J., and Anthony, Q. (2024).
\newblock Zyda: A 1.3 t dataset for open language modeling.
\newblock {\em arXiv preprint arXiv:2406.01981}.

\end{thebibliography}

\newpage

\clearpage
\section*{Appendix}

\subsection{Analysis of Global Duplicates}
\begin{figure}[h]
    \centering
    \begin{minipage}{.32\textwidth}
        \centering
        \includegraphics[width=\textwidth]{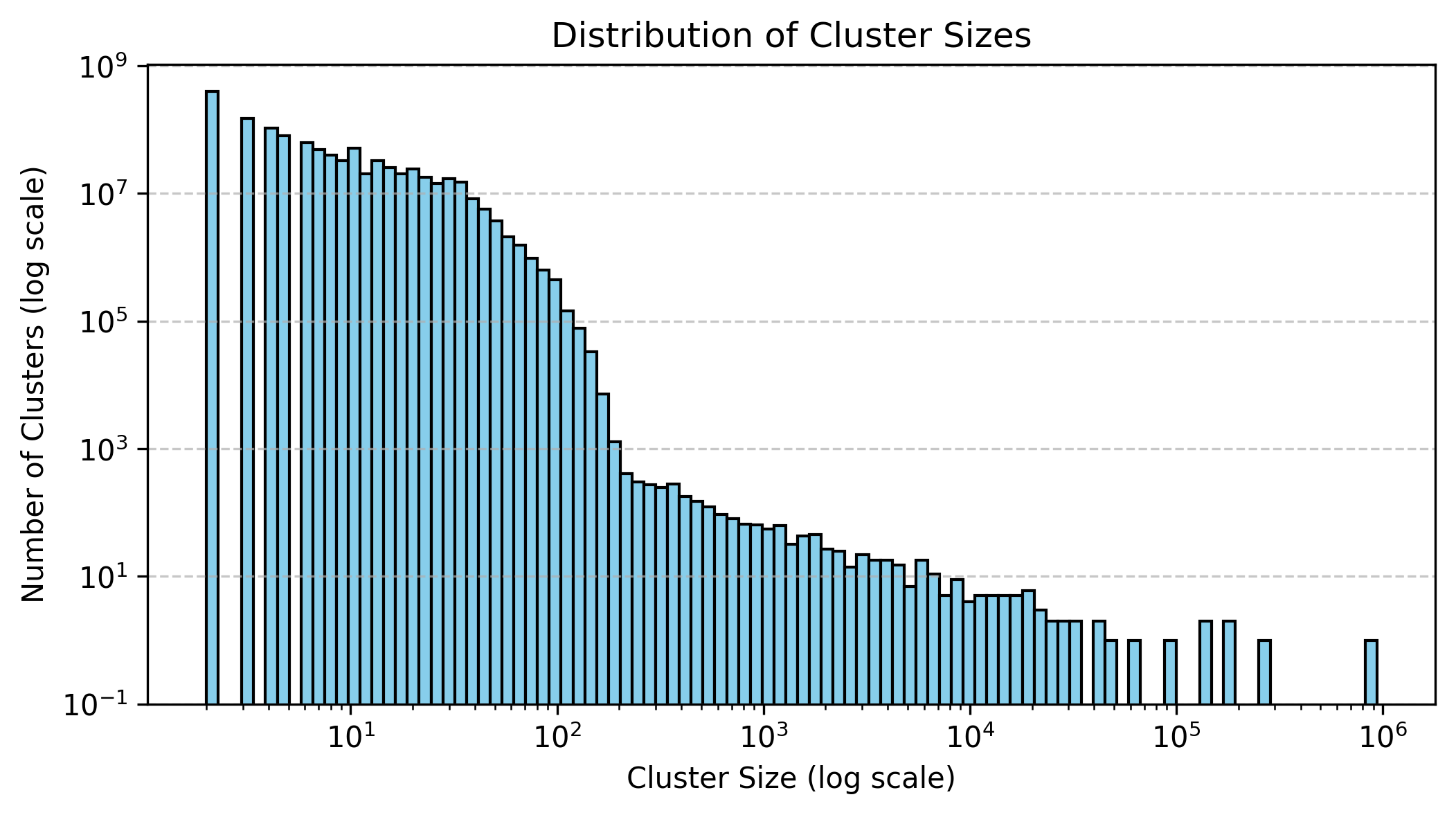}
        \caption{Distribution of cluster sizes of duplicates in global dataset (log-log scale).}
        \label{fig:distr_clusters_overall}
    \end{minipage}\hfill
    \begin{minipage}{.32\textwidth}
        \centering
        \includegraphics[width=\textwidth]{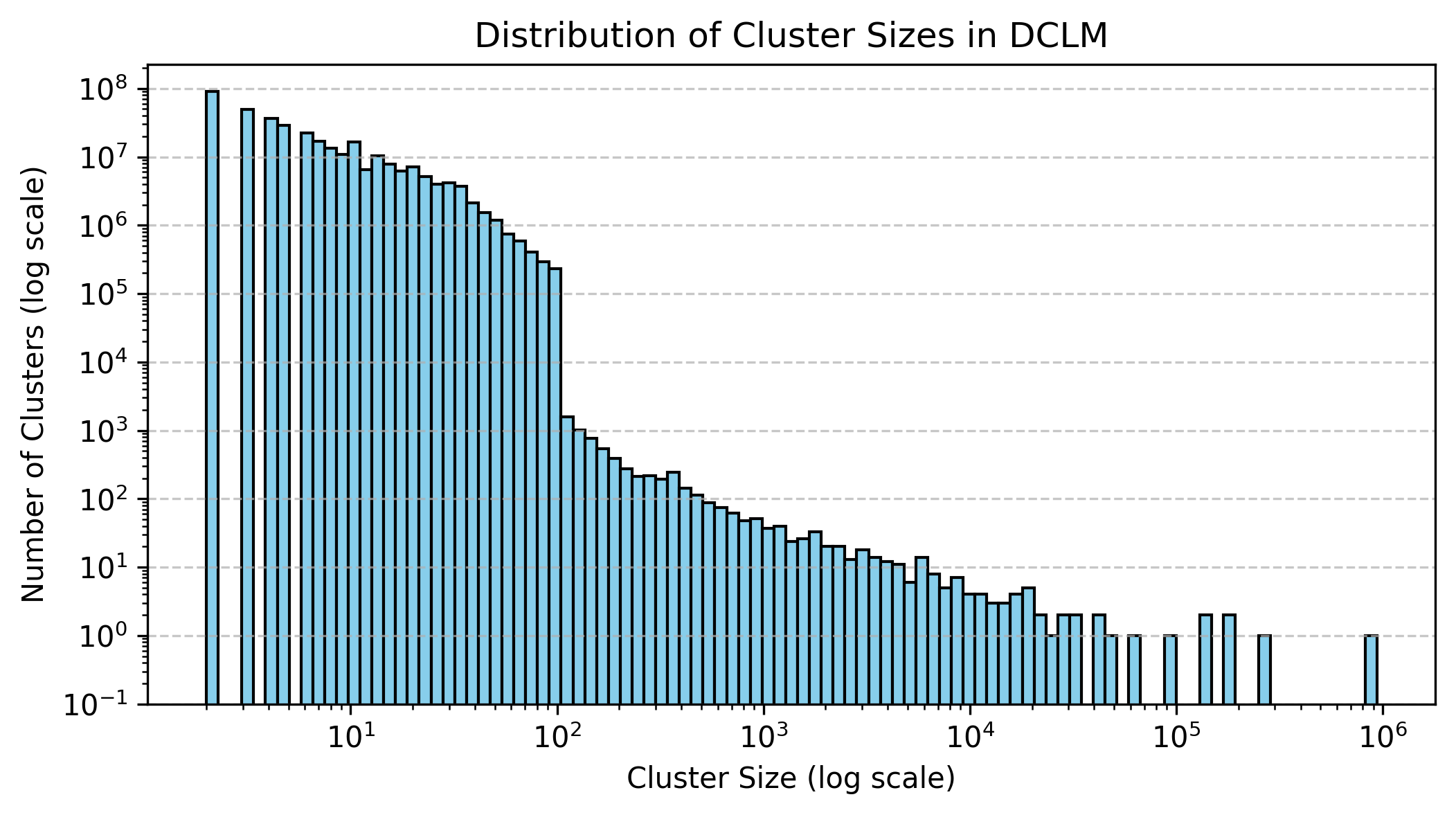}
        \caption{Distribution of cluster sizes of duplicates in DCLM (log-log scale).}
        \label{fig:distr_clusters_dclm}
    \end{minipage}\hfill
    \begin{minipage}{.32\textwidth}
        \centering
        \includegraphics[width=\textwidth]{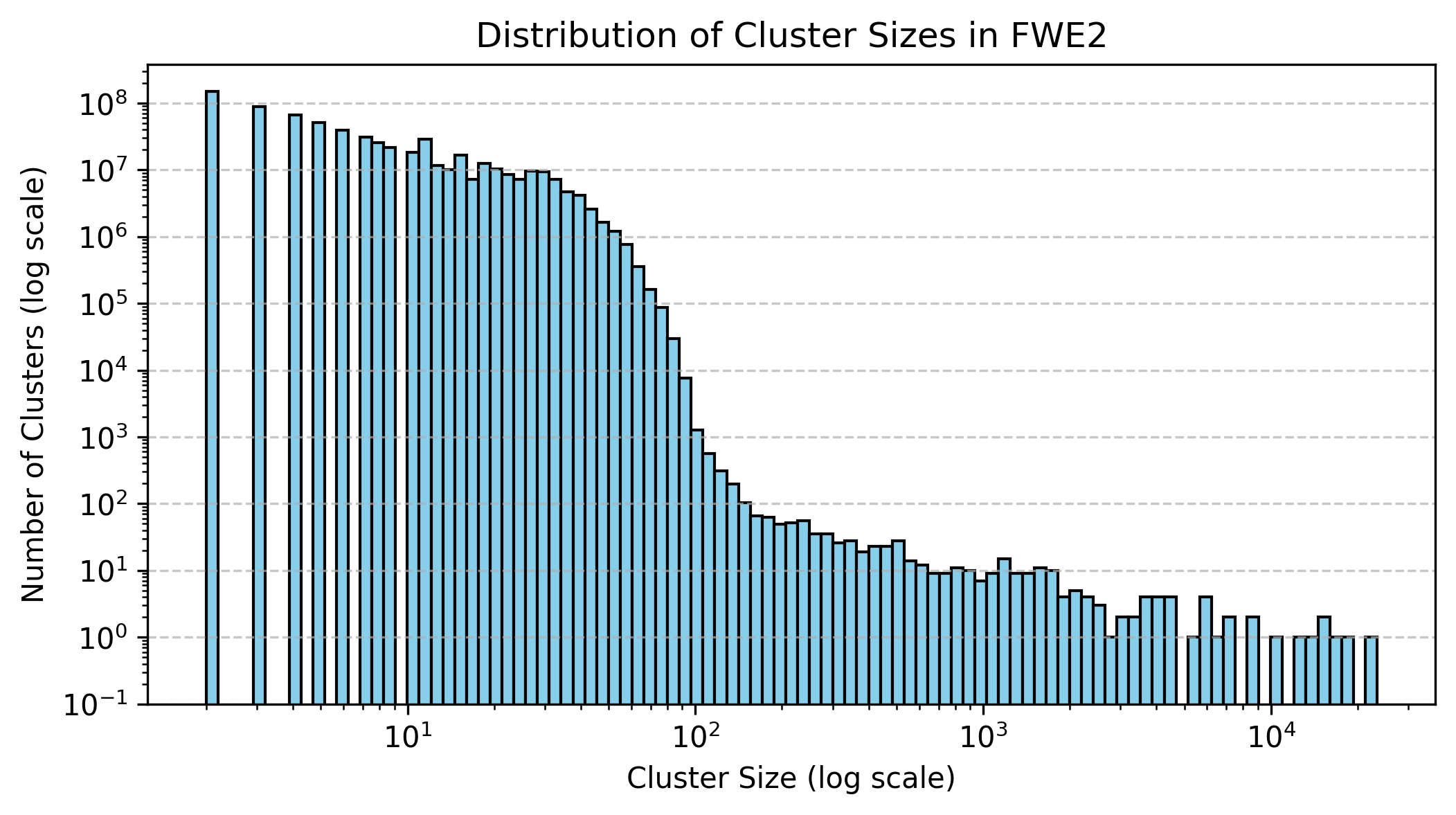}
        \caption{Distribution of cluster sizes of duplicates in FineWeb-Edu2 (log-log scale).}
        \label{fig:distr_clusters_fwe2}
    \end{minipage}
\end{figure}

\begin{figure}[h]
    \centering
    \begin{minipage}{.49\textwidth}
        \centering
        \includegraphics[width=\textwidth]{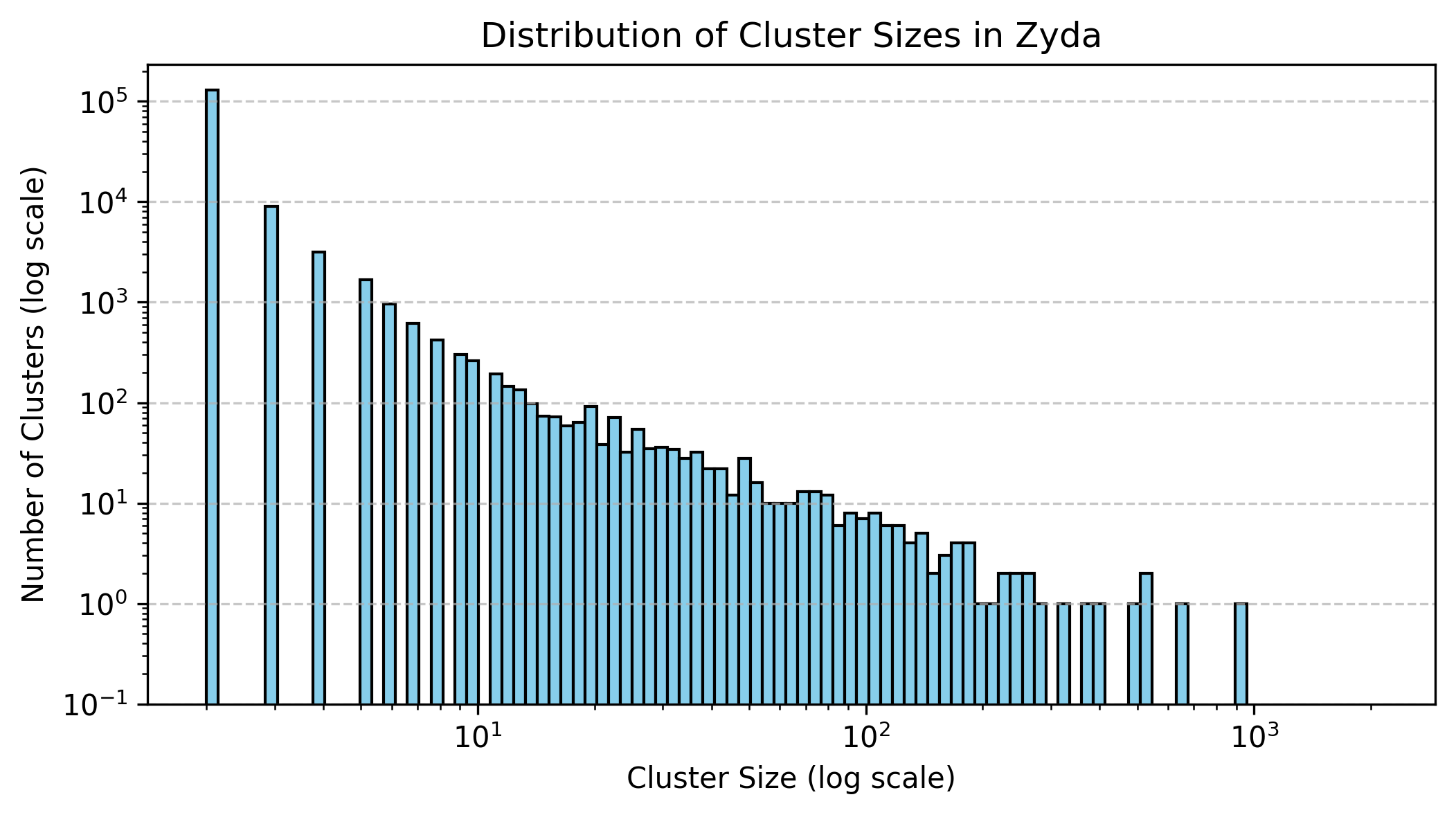}
        \caption{Distribution of cluster sizes of duplicates in Zyda-1 (log-log scale).}
        \label{fig:distr_clusters_zyda}
    \end{minipage}\hfill
    \begin{minipage}{.49\textwidth}
        \centering
        \includegraphics[width=\textwidth]{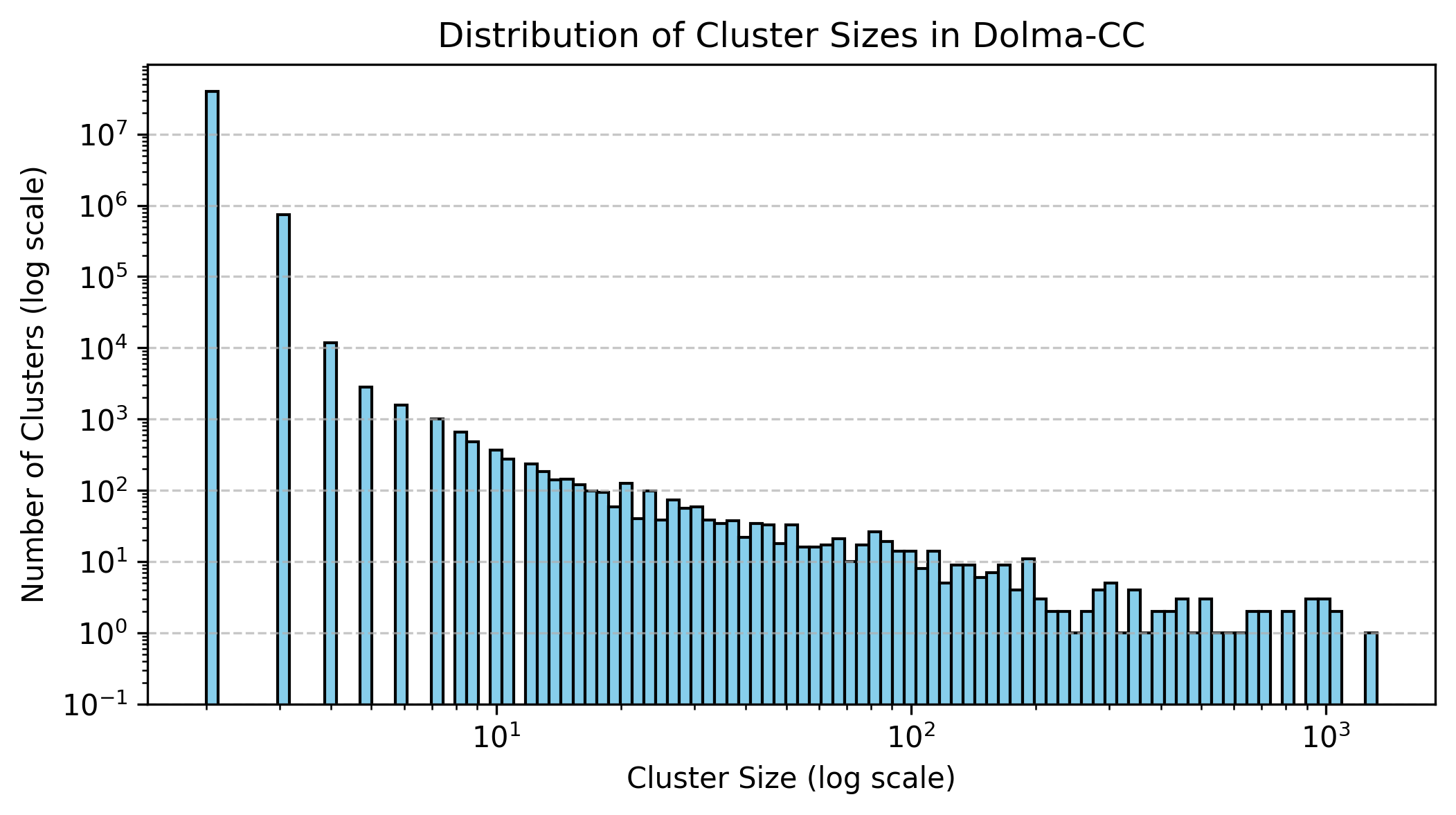}
        \caption{Distribution of cluster sizes of duplicates in Dolma-CC v1.7 (log-log scale).}
        \label{fig:distr_clusters_dolma-cc}
    \end{minipage}
\end{figure}

We present histograms depicting the distribution of duplicate cluster sizes in all the datasets (see Fig.\ref{fig:distr_clusters_overall}-\ref{fig:distr_clusters_dolma-cc}). Please, note that all the figures uselog-log scale. We see a significant drop in the number of duplicate clusters starting from around a size of 100 duplicates. This drop is present both in DCLM and FineWeb-Edu2 (see Fig. \ref{fig:distr_clusters_dclm} and \ref{fig:distr_clusters_fwe2} respectively), and most likely is explained by a combination of the deduplication strategy when creating both datasets: DCLM deduplication was done individually within 10 shards, while FineWeb-Edu2 was deduplicated within every Common Crawl snapshot. We find that large clusters usually contain low quality material (repeated advertisements, license agreements templates, etc), so it’s not surprising that such documents were removed. Notably, DCLM still contained one cluster with the size close to 1 million documents, containing low quality documents seemingly coming from the advertisements, which was somehow accepted by their quality filter.

We find that both Zyda-1 and Dolma-CC contain only a small amount of duplicates, which is expected, since both datasets were deduplicated globally by their authors. Remaining duplicates are likely false negatives from the initial deduplication procedure, or false positives from our own deduplication. Note, that the distribution of duplicate clusters sizes of these two datasets (Fig. \ref{fig:distr_clusters_zyda} and \ref{fig:distr_clusters_dolma-cc}) don’t contain any sharp drops, but rather hyper exponentially decreases with cluster size.

\onecolumn
\twocolumn

\clearpage
\newpage

\subsection{Largest cluster in DCLM}
Below is an example of the document from the largest cluster (roughly 1M documents) of duplicates in DCLM (quality score 0.482627):

\begin{textbox}
Is safe? Is scam?
Is safe for your PC?
Is safe or is it scam?
Domain is Safe
Safe score: 1

The higher the number, the more dangerous the website.
Any number higher than 1 means DANGER.

Positive votes:
Negative votes:
Vote Up Vote Down review

Have you had bad experience with Warn us, please!
\end{textbox}

As can be observed this appears to be some kind of fairly low quality advertisement for some computer security product. 

\subsection{Examples of varying quality score in DCLM in a cluster}

To get a sense of what the DCLM quality filter is doing, we present below a few documents, which are selected from the same duplicate cluster, but with different quality scores from DCLM. Quality score varies from ~0.2 (high quality) to ~0.04 (low quality).

\textbf{Quality score of: 0.19616}
\begin{textbox}
Thrill Jockey instrumental duo Rome are, like many of the acts on the Chicago-based independent label, generally categorized as loose adherents of "post-rock," a period-genre arising in the mid-'90s to refer to rock-based bands utilizing the instruments and structures of music in a non-traditionalist or otherwise heavily mutated fashion. Unlike other Thrill Jockey artists such as Tortoise and Trans-Am, however, Rome draw less obviously from the past, using instruments closely associated with dub (melodica, studio effects), ambient (synthesizers, found sounds), industrial (machine beats, abrasive sounds), and space music (soundtrack-y atmospherics), but fashioning from them a sound which clearly lies beyond the boundaries of each. Perhaps best described as simply "experimental," Rome formed in the early '90s as the trio of Rik Shaw (bass), Le Deuce (electronics), and Elliot Dicks (drums). Based in Chicago, their Thrill Jockey debut was a soupy collage of echoing drums, looping electronics, and deep, droning bass, with an overwhelmingly live feel (the band later divulged that much of the album was the product of studio jamming and leave-the-tape-running-styled improvisation). Benefiting from an early association with labelmates Tortoise as representing a new direction for American rock, Rome toured the U.S. and U.K. with the group (even before the album had been released), also appearing on the German Mille Plateaux label's tribute compilation to French philosopher Gilles Deleuze, In Memoriam. Although drummer Dicks left the group soon after the first album was released, Shaw and Deuce wasted no time with new material, releasing the "Beware Soul Snatchers" single within weeks of its appearance. An even denser slab of inboard studio trickery, "Soul Snatchers" was the clearest example to date of the group's evolving sound, though further recordings failed to materialize. ~ Sean Cooper, Rovi
\end{textbox}

\textbf{Quality score: 0.091928}
\begin{textbox}
Thrill Jockey instrumental duo Rome are, like many of the acts on the Chicago-based independent label, generally grouped in as loose adherents of "post-rock," a period-genre arising in the mid-'90s to refer to rock-based bands utilizing the instruments and structures of the music in a non-traditionalist or otherwise heavily mutated fashion. Unlike other Thrill Jocky artists such as Tortoise and Trans-Am, however, Rome draw less obviously from the past, using instruments closely associated with dub (melodica, studio effects), ambient (synthesizers, found sounds), industrial (machine beats, abrasive sounds), and space music (soundtrack-y atmospherics), but fashioning from them a sound which lay clearly beyond the boundaries of each. Perhaps best described as simply experimental, Rome formed in the early '90s as the trio of Rik Shaw (bass), Le Deuce (electronics), and Elliot Dick (drums). Based in Chicago, their Thrill Jockey debut was a soupy collage of echoing drums, looping electronics, and deep, droning bass, with an overwhelmingly live feel (the band later divulged that much of the album was the product of studio jamming and leave-the-tape-running styled improvisation). Benefiting from an early association with labelmates Tortoise as representing a new direction for American rock, Rome toured the U.S. and U.K. with the group (even before the album had been released), also appearing on the German Mille Plateaux label's tribute compilation to French philosopher Gilles Deleuze, In Memoriam. Although drummer Elliot Dick left the group soon after the first album was released, Shaw and Deuce wasted no time with new material, releasing the "Beware Soul Snatchers" single within weeks of its appearance. An even denser slab of inboard studio trickery, "Soul Snatchers" was the clearest example to date of the group's evolving sound, though further recordings failed to materialize.
Sean Cooper, Rovi

More Rome

You may also like...
\end{textbox}

\textbf{Quality score: 0.072259}
\begin{textbox}
recent on-air advertisers

Now Playing

You Control the ...

Artist Snapshot:

    Thrill Jockey instrumental duo Rome are, like many of the acts on the Chicago-based independent label, generally grouped in as loose adherents of "post-rock," a period-genre arising in the mid-'90s to refer to rock-based bands utilizing the instruments and structures of the music in a non-traditionalist or otherwise heavily mutated fashion. Unlike other Thrill Jocky artists such as Tortoise and Trans-Am, however, Rome draw less obviously from the past, using instruments closely associated with dub (melodica, studio effects), ambient (synthesizers, found sounds), industrial (machine beats, abrasive sounds), and space music (soundtrack-y atmospherics), but fashioning from them a sound which lay clearly beyond the boundaries of each. Perhaps best described as simply experimental, Rome formed in the early '90s as the trio of Rik Shaw (bass), Le Deuce (electronics), and Elliot Dick (drums). Based in Chicago, their Thrill Jockey debut was a soupy collage of echoing drums, looping electronics, and deep, droning bass, with an overwhelmingly live feel (the band later divulged that much of the album was the product of studio jamming and leave-the-tape-running styled improvisation). Benefiting from an early association with labelmates Tortoise as representing a new direction for American rock, Rome toured the U.S. and U.K. with the group (even before the album had been released), also appearing on the German Mille Plateaux label's tribute compilation to French philosopher Gilles Deleuze, In Memoriam. Although drummer Elliot Dick left the group soon after the first album was released, Shaw and Deuce wasted no time with new material, releasing the "Beware Soul Snatchers" single within weeks of its appearance. An even denser slab of inboard studio trickery, "Soul Snatchers" was the clearest example to date of the group's evolving sound, though further recordings failed to materialize. ~ Sean Cooper, Rovi
\end{textbox}

\textbf{Quality score: 0.0424}
\begin{textbox}

18 June 2015

ROME self titled 1996

by request

Artist Biography by

Thrill Jockey instrumental duo Rome are, like many of the acts on the Chicago-based independent label, generally categorized as loose adherents of "post-rock," a period-genre arising in the mid-'90s to refer to rock-based bands utilizing the instruments and structures of music in a non-traditionalist or otherwise heavily mutated fashion. Unlike other Thrill Jockey artists such as Tortoise and Trans-Am, however, Rome draw less obviously from the past, using instruments closely associated with dub (melodica, studio effects), ambient (synthesizers, found sounds), industrial (machine beats, abrasive sounds), and space music (soundtrack-y atmospherics), but fashioning from them a sound which clearly lies beyond the boundaries of each. Perhaps best described as simply "experimental," Rome formed in the early '90s as the trio of Rik Shaw (bass), Le Deuce (electronics), and Elliot Dicks (drums). Based in Chicago, their Thrill Jockey debut was a soupy collage of echoing drums, looping electronics, and deep, droning bass, with an overwhelmingly live feel (the band later divulged that much of the album was the product of studio jamming and leave-the-tape-running-styled improvisation). Benefiting from an early association with labelmates Tortoise as representing a new direction for American rock, Rome toured the U.S. and U.K. with the group (even before the album had been released), also appearing on the German Mille Plateaux label's tribute compilation to French philosopher Gilles Deleuze, In Memoriam. Although drummer Dicks left the group soon after the first album was released, Shaw and Deuce wasted no time with new material, releasing the "Beware Soul Snatchers" single within weeks of its appearance. An even denser slab of inboard studio trickery, "Soul Snatchers" was the clearest example to date of the group's evolving sound, though further recordings failed to materialize.

1 Leaving Perdition 8:10
2 Intermodal 3:39
3 Lunar White 3:25
4 She's A Black Belt 3:14
5 Rohm 1:09
6 Radiolucence (Version) 5:31
7 Deepest Laws 14:14

No comments:
\end{textbox}

\end{document}